\documentclass[10pt,twocolumn,letterpaper]{article}

\usepackage[pagenumbers]{cvpr}  

\ifdefined\pdfminorversion
\fi

\usepackage{array}
\usepackage{multirow}
\usepackage{float}
\usepackage{amsthm}
\usepackage{algorithm}
\usepackage{algorithmic}
\usepackage{tikz}

\newcolumntype{P}[1]{>{\raggedright\arraybackslash}p{#1}}

\definecolor{cvprblue}{rgb}{0.21,0.49,0.74}
\usepackage[breaklinks,colorlinks,allcolors=cvprblue]{hyperref}
\def\confName{CVPR}
\def\confYear{2027}

\title{On Demand, Not Uniformity: Adaptive Expert Orchestration for \\  Multimodal Continual Instruction Tuning}

\author{%
Mengxin Qin \quad Kun Wei\thanks{Corresponding author.} \quad Weitao An \quad Jiahua Li \quad Xu Yang \quad Cheng Deng\\
School of Electronic Engineering, Xidian University, Xi'an 710071, China
}

\begin{document}
\maketitle
\begin{abstract}
Multimodal Continual Instruction Tuning (MCIT) enables Multimodal Large Language Models (MLLMs) to adapt to sequential instruction data while retaining previously acquired knowledge. 
While recent Mixture-of-Experts (MoE)-based methods mitigate forgetting by freezing historical experts and introducing new ones, they largely overlook the varying adaptation demands across tasks. 
Probing incoming tasks reveals pronounced heterogeneity in layer-wise gradient pressure and token-wise prediction difficulty. 
This observation highlights two unresolved decisions: where additional capacity is needed and how strongly new experts should contribute to individual tokens.
To address both questions, we propose the Coverage-Calibrated Expert Injection and Residual Arbitration framework (\textbf{CIRA-MoE}), which derives expert acceptance margins via learnable threshold routing to serve as a unified coverage signal. 
Specifically, historical acceptance evidence guides the selective injection of new experts into insufficiently covered layers. Meanwhile, relative evidence between historical and current experts modulates token-wise residual contributions, coupled with arbitration-consistent regularization to promote expert specialization.
Comprehensive experiments on UCIT and CoIN demonstrate that CIRA-MoE outperforms existing state-of-the-art methods in MCIT.
\end{abstract}

\section{Introduction}
\label{sec:intro}

\begin{figure}[t]
    \centering
    \includegraphics[width=0.95\linewidth]{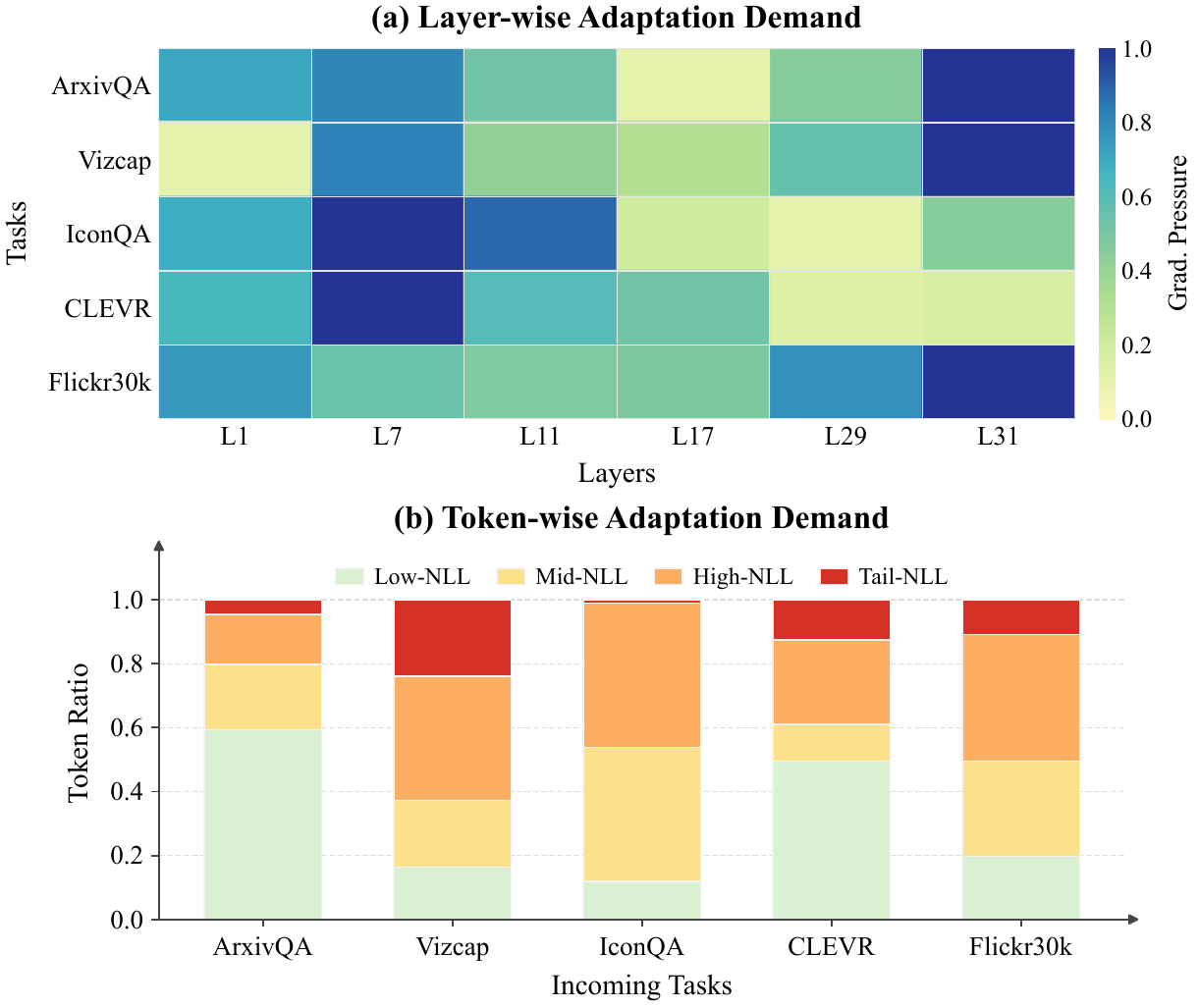}
    \caption{
    Layer- and token-wise heterogeneity in MCIT.
    (a) Layer-wise demand measured by gradient pressure. (b) Token-wise demand characterized by the distribution of teacher-forced NLL.
    }
    \label{fig:motivation}
\end{figure}

Multimodal Large Language Models (MLLMs)~\cite{liu2023visual}
show strong performance on many vision--language tasks.
However, real-world applications require MLLMs to continually
adapt to new visual domains, tasks, and instruction formats.
Multimodal Continual Instruction Tuning (MCIT)~\cite{chen2024coin,guo2025hide}
addresses this challenge by enabling MLLMs to acquire new
capabilities while retaining previously learned knowledge.
Despite recent progress, MCIT remains susceptible to
catastrophic forgetting~\cite{kirkpatrick2017overcoming}.

Existing MCIT methods mainly rely on rehearsal, regularization, or parameter-efficient tuning~\cite{lei2023symbolic,kirkpatrick2017overcoming,zeng2025modalprompt}.
Within parameter-efficient tuning, Mixture-of-Experts (MoE) architectures offer a promising paradigm for continual learning by organizing lightweight modules into expandable expert pools~\cite{chen2024coin,huai2025cl,xie2026same,ge2025dynamic}.
By freezing historical experts and incrementally introducing new ones for incoming tasks, this modular structure enables flexible capacity expansion and parameter isolation to mitigate catastrophic forgetting.

However, incremental MoE methods typically expand capacity and route tokens uniformly, ignoring the varying adaptation demands of incoming tasks.
Recent studies mitigate router and expert drift by regulating parameter updates (SAME~\cite{xie2026same}), guiding layer-wise expansion via task probes (LLaVA-CMoE~\cite{zhao2025llava}), or regularizing token-level routing scores against newly added experts (LLaVA-DyMoE~\cite{zhao2026token}).
Nevertheless, these methods still lack an intrinsic mechanism to identify where frozen historical experts fall short and where new-expert intervention is genuinely needed.

To examine where frozen historical experts fall short across granularities, we probe a subset of task $T_t$'s training data using the model trained through task $T_{t-1}$ before adaptation.
\textbf{(1) At the layer level}, we measure normalized
gradient pressure at candidate expert layers.
As shown in Fig.~\ref{fig:motivation}(a), adaptation demand
varies substantially across layers and tasks.
For example, ArxivQA exhibits strong gradient pressure
at L31 but low pressure at L17, whereas IconQA shows
high pressure at L7 and L11.
\textbf{(2) At the token level}, we use teacher-forced
NLL on target response tokens as a proxy for the historical
model's prediction difficulty.
As shown in Fig.~\ref{fig:motivation}(b), many target
tokens receive low NLL, whereas only a subset exhibits
high prediction difficulty, with its proportion varying
across tasks.
These observations reveal heterogeneous adaptation demands across both layers and tokens.
This heterogeneity highlights two critical decisions for continual expert expansion: where additional capacity is needed and how strongly new experts should intervene on individual tokens.

To answer both questions, we propose the \textbf{C}overage-Calibrated Expert \textbf{I}njection and \textbf{R}esidual \textbf{A}rbitration framework (\textbf{CIRA-MoE}), a demand-aligned adaptation approach that coordinates expert allocation and residual arbitration across layers and tokens.
Specifically, CIRA-MoE employs Expert-Wise Acceptance Routing (EWAR) to assign each expert an independently learnable acceptance threshold, yielding signed acceptance margins as a unified coverage signal.
Building on this signal, Layer-Wise Expert Injection (LWEI) aggregates historical acceptance evidence to quantify layer-wise coverage gaps and selectively injects new experts into insufficiently covered layers.
Meanwhile, Token-Wise Residual Arbitration (TWRA) compares historical and current expert evidence to continuously modulate current-expert residual contributions for individual tokens.
We further introduce Arbitration-Consistent Regularization (ACR) to suppress unnecessary current-expert activation and encourage expert specialization, aligning expert utilization with the arbitration decisions.

Our main contributions are summarized as follows:
\begin{itemize}

\item  We reveal the layer- and token-wise heterogeneity of adaptation demand in MoE-based MCIT and demonstrate that threshold-based historical acceptance margins faithfully capture these coverage gaps.

\item We propose \textbf{CIRA-MoE}, which coordinates LWEI and TWRA
using expert acceptance evidence and employs ACR to regularize
expert utilization.

\item Comprehensive experiments on UCIT and CoIN demonstrate that CIRA-MoE outperforms existing state-of-the-art methods in MCIT.

\end{itemize}


\section{Related Work}
\label{sec:related}
\subsection{Multimodal Continual Instruction Tuning}
MCIT aims to continually adapt MLLMs to sequential tasks while preserving previously acquired knowledge~\cite{chen2024coin}. 
Existing approaches mainly mitigate catastrophic forgetting through regularization, rehearsal, and parameter-efficient adaptation. 
Regularization-based methods constrain parameter updates or model outputs via weight consolidation, knowledge distillation, or gradient projection~\cite{kirkpatrick2017overcoming,li2016learning,chen2025sefe,li2026multimodal,liu2026octopus}. 
Rehearsal-based methods replay stored or reconstructed samples, such as scene-graph- and question-based replay~\cite{lei2023symbolic,marouf2025ask}. 
Parameter-efficient approaches instead isolate task-specific knowledge using lightweight modules, such as learnable prompts~\cite{wang2022learning,zeng2025modalprompt} and orthogonal or expandable low-rank adapters~\cite{wang2023orthogonal}.

\subsection{LoRA and MoE-Based Adaptation}
Low-Rank Adaptation (LoRA) represents task-specific updates with low-rank matrices~\cite{hu2022lora} and serves as a common building block for continual adaptation. 
For instance, HiDe-LLaVA employs hierarchical LoRA decoupling to reduce cross-task interference, whereas ProgLoRA progressively expands LoRA modules across tasks~\cite{guo2025hide,yu2025progressive}. 
Building on LoRA, MoE-based methods organize lightweight adapters into routed expert pools to enhance capacity and modularity, as exemplified by MoELoRA and CL-MoE~\cite{chen2024coin,huai2025cl}. 
Recent work such as CoRe-MoE further promotes expert reuse by constructing compact coordinate experts from shared low-rank directions~\cite{liu-etal-2026-core}.

\subsection{Expert Evolution and Routing Stability}
To move beyond uniform expert allocation and static top-$k$ routing, recent studies investigate adaptive expert allocation and routing stabilization. Along the structural dimension, D-MoLE allocates LoRA experts across layers under a parameter budget~\cite{ge2025dynamic}, probe-guided approaches select expansion locations and task-specific routing components~\cite{zhao2025llava,wang2026cope}, and CRAM combines adaptive-rank allocation with centroid-guided routing~\cite{tang2026cram}. Along the routing dimension, SAME regulates routing directions and expert updates~\cite{xie2026same}, LLaVA-DyMoE alleviates token-level routing drift~\cite{zhao2026token}, and PASs-MoE mitigates misaligned co-drift between routers and experts via pathway activation subspaces~\cite{hou-etal-2026-pass}. 
However, existing approaches do not explicitly quantify the coverage of frozen historical experts, making it difficult to determine where prior knowledge falls short and where new experts need to intervene. 
In contrast, CIRA-MoE introduces threshold-based acceptance margins as a unified coverage signal to coordinate layer-wise expert injection and token-wise residual arbitration on demand.

\section{Method}
\begin{figure*}[t]
    \centering
    \includegraphics[width=0.9\linewidth]{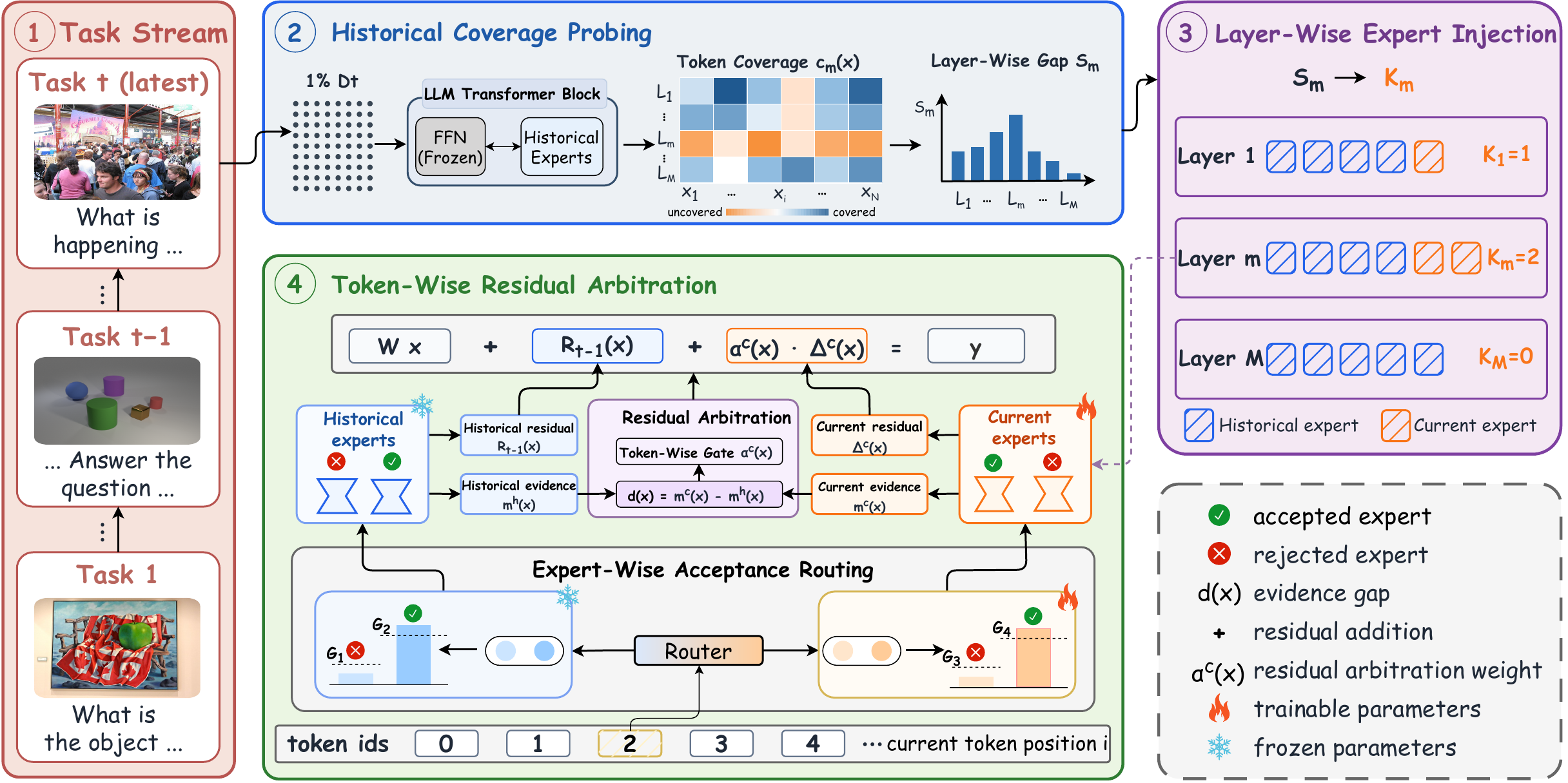}
\caption{
Overview of CIRA-MoE.
For each incoming multimodal instruction task, CIRA-MoE:
(1) receives the latest task;
(2) probes frozen historical experts using a 1\% subset to estimate historical coverage $c_m(\mathbf{x})$ and layer-wise coverage gaps $S_m$;
(3) injects up to two current experts per FFN projection according to $S_m$, allocating more capacity to poorly covered positions; and
(4) applies Expert-Wise Acceptance Routing and Token-Wise Residual Arbitration to select experts and regulate current residual contributions.
After training, current experts are frozen and retained with their stage indices; the same residual aggregation is used at inference.
}
\label{fig:framework}
\end{figure*}

\subsection{Problem Formulation}

We study multimodal continual instruction tuning over a task sequence $\mathcal{T}=\{T_1,T_2,\ldots,T_N\}$. At stage $t$, the model only observes the current training set $\mathcal{D}_t=\{(\mathbf{x}^{v},\mathbf{x}^{q},\mathbf{x}^{a})\}$, where $\mathbf{x}^{v}$, $\mathbf{x}^{q}$, and $\mathbf{x}^{a}$ denote the visual input, textual instruction, and target response. The goal is to adapt to $T_t$ while preserving capabilities learned from $T_1,\ldots,T_{t-1}$.

As shown in Fig.~\ref{fig:framework}, CIRA-MoE uses
Expert-Wise Acceptance Routing (EWAR) to obtain expert-specific
acceptance margins.
Based on this evidence, Layer-Wise Expert Injection (LWEI) allocates new experts according to historical coverage gaps, while Token-Wise Residual Arbitration (TWRA) compares
historical and current acceptance evidence to modulate current-expert residual contributions for each token.
Arbitration-Consistent Regularization (ACR) aligns
current-expert utilization with these arbitration decisions
during training.

\subsection{Expert-Wise Acceptance Routing}
\label{sec:acceptance-routing}
At each FFN linear projection $m$, EWAR applies threshold-based
Top-Any gating~\cite{guo2025dynamic} to an independently allocated
expert pool. Each token activates a variable number of experts;
their acceptance margins guide LWEI and TWRA.
Position indices are omitted below.

Given expert-specific routing prototypes $\mathbf{W}_g=[\mathbf{w}_{g,1},\ldots,\mathbf{w}_{g,K}]\in\mathbb{R}^{d\times K}$ and a token representation $\mathbf{x}\in\mathbb{R}^{d}$, the cosine similarity for expert $e$ is
\begin{equation}
s_e(\mathbf{x})=
\frac{\langle\mathbf{x},\mathbf{w}_{g,e}\rangle}
{\|\mathbf{x}\|\,\|\mathbf{w}_{g,e}\|}.
\end{equation}
Each expert has a learnable threshold $G_e$. Its acceptance margin is defined as
\begin{equation}
u_e(\mathbf{x})=
\sigma(s_e(\mathbf{x}))-\sigma(G_e).
\label{eq:acceptance-margin}
\end{equation}
Here $\sigma$ is the logistic sigmoid. The accepted expert set is
\begin{equation}
\mathcal{A}(\mathbf{x})=
\{e\mid u_e(\mathbf{x})>0\}.
\label{eq:activated-set}
\end{equation}

To enable gradient-based optimization through hard expert selection, we apply an identity straight-through estimator (STE) to the binary gate $g_e(\mathbf{x})=\mathbb{I}[u_e(\mathbf{x})>0]$, using $\partial g_e/\partial u_e\approx 1$ in the backward pass to propagate gradients through the continuous margin $u_e(\mathbf{x})$.

\subsection{Layer-Wise Expert Injection}
\label{sec:coverage-injection}

EWAR enables flexible reuse of historical experts but cannot determine where an incoming task requires additional capacity. LWEI addresses this limitation by measuring coverage gaps in frozen historical experts before training.

For the first task, where no historical experts are available, we skip coverage probing and initialize a seed pool with $K_m^0=K_{\mathrm{seed}}$ experts at each candidate position $m$, where $K_{\mathrm{seed}}=1$ in our experiments.

For each subsequent task $T_t$, we sample a probing subset $\mathcal{D}_t^p\subset\mathcal{D}_t$ and evaluate historical coverage at each candidate position $m$. Using the acceptance margin in Eq.~\eqref{eq:acceptance-margin}, we define the best historical acceptance margin as
\begin{equation}
c_m(\mathbf{x})=
\max_{e\in\mathcal{E}_m^h}u_e(\mathbf{x}),
\end{equation}
where $\mathcal{E}_m^h$ denotes the historical expert pool at position $m$. A positive $c_m(\mathbf{x})$ indicates that at least one historical expert accepts the token, whereas a non-positive value indicates that all historical experts reject it.

Let $\Omega_t^p$ denote the valid token positions in $\mathcal{D}_t^p$, and let $\mathbf{x}_{m,i}$ be the representation at position $m$. With $[z]_+=\max(0,z)$, we quantify the layer-wise coverage gap as
\begin{equation}
S_m=
\frac{1}{|\Omega_t^p|}
\sum_{i\in\Omega_t^p}
\left[-c_m(\mathbf{x}_{m,i})\right]_+.
\end{equation}
A larger $S_m$ indicates weaker historical coverage and greater demand for additional capacity.

Let $\mathcal{M}_+=\{m\mid S_m>0\}$ denote the positions with positive coverage gaps. If $\mathcal{M}_+$ is nonempty, we compute their average gap:
\begin{equation}
\bar{S}_+=
\frac{1}{|\mathcal{M}_+|}
\sum_{j\in\mathcal{M}_+}S_j.
\end{equation}

The number of new experts injected at position $m$ is
\begin{equation}
K_m^c=
\left\lceil
K_{\max}\frac{S_m}{S_m+\bar{S}_+}
\right\rceil
\quad \text{for } m\in\mathcal{M}_+,
\end{equation}
and $K_m^c=0$ otherwise, where $K_{\max}=2$ bounds new experts per position.

\subsection{Token-Wise Residual Arbitration}
\label{sec:residual-arbitration}

After LWEI, experts are organized into task-specific groups,
containing frozen historical experts and trainable current experts.
TWRA uses their relative acceptance evidence to modulate
the contribution of each expert group's residual.

For task group $j$, let $\mathcal{E}^{(j)}$ denote its experts
and $\mathcal{A}^{(j)}(\mathbf{x})=
\mathcal{A}(\mathbf{x})\cap\mathcal{E}^{(j)}$
the accepted subset for token $\mathbf{x}$.
Let $E_e(\mathbf{x})$ denote expert $e$'s LoRA residual output.
The group's mean residual and maximum acceptance evidence are
\begin{equation}
\begin{aligned}
\Delta^{(j)}(\mathbf{x})
&=
\frac{1}{|\mathcal{A}^{(j)}(\mathbf{x})|}
\sum_{e\in\mathcal{A}^{(j)}(\mathbf{x})}
g_e(\mathbf{x})E_e(\mathbf{x}),\\
m^{(j)}(\mathbf{x})
&=
\max_{e\in\mathcal{A}^{(j)}(\mathbf{x})}
u_e(\mathbf{x}),
\end{aligned}
\label{eq:group-evidence}
\end{equation}
where both quantities are zero if $\mathcal{A}^{(j)}(\mathbf{x})=\emptyset$.

For the first task, we set $\alpha^{(1)}(\mathbf{x})=1$.
For each subsequent task group $j>1$, its relative evidence
gap is
\begin{equation}
d^{(j)}(\mathbf{x})
=
m^{(j)}(\mathbf{x})
-
\max_{i<j}m^{(i)}(\mathbf{x}).
\end{equation}
The corresponding arbitration weight is
\begin{equation}
\alpha^{(j)}(\mathbf{x})
=
\sigma\left(\frac{d^{(j)}(\mathbf{x})}{\tau}\right)
\cdot
\left[
1-\exp\left(-\frac{|d^{(j)}(\mathbf{x})|}{\tau}\right)
\right],
\label{eq:twra-weight}
\end{equation}
where $\tau$ controls sensitivity to the evidence gap.
This rule favors groups with stronger acceptance
evidence, suppresses contributions near equal
evidence, and continuously attenuates groups with
weaker evidence.

The aggregated residual of the first $t$ task groups is
\begin{equation}
R_t(\mathbf{x})
=
\sum_{j=1}^{t}
\alpha^{(j)}(\mathbf{x})\Delta^{(j)}(\mathbf{x}),
\qquad R_0(\mathbf{x})=0.
\label{eq:historical-residual}
\end{equation}
The adapted projection outputs $\mathbf{y}=W\mathbf{x}+R_t(\mathbf{x})$,
where $W$ is the frozen projection matrix.

\paragraph{Training and inference.}
At training stage $t$, groups $j<t$ are frozen historical
experts, while group $t$ (denoted by superscript $c$ for current) contains trainable experts.
The forward residual is
\begin{equation}
\Delta_{\mathrm{train}}(\mathbf{x})
=
R_{t-1}(\mathbf{x})
+
\operatorname{sg}[\alpha^{c}(\mathbf{x})]
\Delta^{c}(\mathbf{x}),
\label{eq:training-residual}
\end{equation}
where $\operatorname{sg}[\cdot]$ stops gradients through
$\alpha^{c}(\mathbf{x})\equiv\alpha^{(t)}(\mathbf{x})$.
After training, group $t$ and its routing parameters are frozen.
At inference, task groups are evaluated in introduction order
via Eqs.~\eqref{eq:group-evidence}--\eqref{eq:historical-residual}
to yield $\Delta_{\mathrm{infer}}(\mathbf{x})=R_t(\mathbf{x})$.
Task indices serve only as internal metadata for
grouping experts; no task identifier is required at inference.

\begin{table*}[t]
\centering
\caption{Performance comparison on UCIT. Avg averages each task over the stages
from its introduction onward; Last reports final task scores.
Bold and underlined values denote the best and second-best continual-learning results within each block.}
\label{tab:ucit_results}
\small
\setlength{\tabcolsep}{3pt}
\renewcommand{\arraystretch}{1.04}
\begin{tabular*}{\textwidth}{@{\extracolsep{\fill}}cl*{7}{r}@{}}
\toprule
 & \textbf{Method} & \textbf{ImgNet-R} & \textbf{ArxivQA}
 & \textbf{Vizcap} & \textbf{IconQA} & \textbf{CLEVR}
 & \textbf{Flickr30k} & \textbf{Average}$\uparrow$ \\
\midrule
 & Zero-shot & 16.27 & 53.73 & 38.39 & 19.20 & 20.63 & 41.88 & 31.68 \\
\midrule
\multirow{8}{*}{\rotatebox{90}{\textbf{Avg}}} & FineTune~\cite{hu2022lora} & 49.31 & 78.40 & 50.48 & 53.44 & 55.53 & \underline{57.95} & 57.52 \\
 & LwF~\cite{li2016learning} & 55.60 & 79.86 & 53.23 & 54.87 & 56.51 & 56.34 & 59.40 \\
 & EWC~\cite{kirkpatrick2017overcoming} & 54.23 & 80.13 & 53.14 & 55.06 & 57.52 & 55.94 & 59.34 \\
 & L2P~\cite{wang2022learning} & 41.52 & 82.32 & 51.98 & 52.21 & 43.16 & 52.77 & 53.99 \\
 & O-LoRA~\cite{wang2023orthogonal} & 75.26 & 86.73 & \textbf{55.86} & 58.47 & 57.38 & 53.52 & 64.54 \\
 & MoELoRA~\cite{chen2024coin} & 64.49 & 82.42 & 49.54 & 56.87 & 56.35 & \textbf{58.34} & 61.33 \\
 & HiDe-LLaVA~\cite{guo2025hide} & \underline{85.70} & \textbf{92.70} & \underline{54.10} & \underline{66.87} & \underline{59.12} & 55.15 & \underline{68.94} \\
\cmidrule(l){2-9}
 & \textbf{CIRA-MoE} & \textbf{86.27} & \underline{89.73} & 51.45 & \textbf{76.37} & \textbf{69.18} & 57.07 & \textbf{71.68} \\
\midrule
\multirow{10}{*}{\rotatebox{90}{\textbf{Last}}} & FineTune~\cite{hu2022lora} & 37.63 & 72.33 & 43.47 & 41.70 & 35.63 & \underline{57.95} & 48.12 \\
 & LwF~\cite{li2016learning} & 40.27 & 75.93 & 42.76 & 44.38 & 37.43 & 56.34 & 49.52 \\
 & EWC~\cite{kirkpatrick2017overcoming} & 39.05 & 77.88 & 43.24 & 45.33 & 39.72 & 55.94 & 50.20 \\
 & L2P~\cite{wang2022learning} & 32.73 & 80.41 & 43.72 & 42.16 & 39.25 & 52.77 & 48.51 \\
 & O-LoRA~\cite{wang2023orthogonal} & 69.36 & 82.42 & \underline{48.64} & 53.66 & 42.53 & 53.52 & 58.36 \\
 & MoELoRA~\cite{chen2024coin} & 49.87 & 77.63 & 43.65 & 46.40 & 36.47 & \textbf{58.34} & 52.06 \\
 & HiDe-LLaVA~\cite{guo2025hide} & 80.50 & \underline{89.83} & \textbf{48.78} & 62.90 & 47.97 & 55.15 & 64.19 \\
 & CL-MoE~\cite{huai2025cl} & 66.33 & 77.00 & 44.78 & 51.87 & 53.53 & 57.42 & 58.49 \\
 & SEFE~\cite{chen2025sefe} & \underline{80.83} & 78.00 & 47.01 & \underline{69.63} & \textbf{65.83} & 57.92 & \underline{66.54} \\
\cmidrule(l){2-9}
 & \textbf{CIRA-MoE} & \textbf{82.10} & \textbf{93.93} & 48.46 & \textbf{70.10} & \underline{65.73} & 57.07 & \textbf{69.56} \\
\bottomrule
\end{tabular*}
\vspace{3pt}
\end{table*}

\subsection{Arbitration-Consistent Regularization}
\label{sec:acr}

We introduce Arbitration-Consistent Regularization (ACR)
to encourage minimal intervention and expert specialization
through two complementary auxiliary losses.

We first define the average current-expert evidence as
\begin{equation}
\bar{\rho}^{c}(\mathbf{x})
=
\frac{1}{|\mathcal{E}^{c}|}
\sum_{e\in\mathcal{E}^{c}}\rho_e(\mathbf{x}),
\end{equation}
where $\rho_e(\mathbf{x})=[u_e(\mathbf{x})]_+$ is the
nonnegative acceptance evidence, so rejected experts
contribute zero. We detach the arbitration weight as
$\tilde{\alpha}^{c}(\mathbf{x})
=\operatorname{sg}[\alpha^{c}(\mathbf{x})]$.

The minimal-intervention loss suppresses unnecessary
current-expert activation when the arbitration weight
is small:
\begin{equation}
\mathcal{L}_{\mathrm{mi}}(\mathbf{x})
=
(1-\tilde{\alpha}^{c}(\mathbf{x}))
\bar{\rho}^{c}(\mathbf{x}).
\end{equation}

The specialization loss encourages a dominant current
expert when current-expert intervention is favored.
For tokens with at least one accepted current expert
and more than one available current expert, we define
\begin{equation}
\mathcal{L}_{\mathrm{sp}}(\mathbf{x})
=
\tilde{\alpha}^{c}(\mathbf{x})
\left[
\delta+\bar{\rho}^{c}(\mathbf{x})
-m^{c}(\mathbf{x})
\right]_+,
\end{equation}
where $m^{c}(\mathbf{x})$ denotes the maximum acceptance
evidence among accepted current experts.
The specialization loss is set to zero otherwise.
The margin $\delta$ controls the required evidence gap
between the dominant expert and the average current-expert
evidence.

The overall objective is
\begin{equation}
\mathcal{L}
=
\mathcal{L}_{\mathrm{task}}
+
\lambda\mathcal{L}_{\mathrm{ACR}},
\end{equation}
where $\mathcal{L}_{\mathrm{task}}$ is autoregressive cross-entropy
over valid answer tokens. $\mathcal{L}_{\mathrm{ACR}}$ averages the
sum of the two auxiliary losses over valid tokens at positions with
nonempty current-expert pools, and $\lambda$ controls its strength.

\section{Experiments}
\subsection{Implementation Details}
\paragraph{Datasets.}
We evaluate on UCIT~\cite{guo2025hide} and CoIN~\cite{chen2024coin}.
UCIT comprises six sequential tasks: ImageNet-R~\cite{hendrycks2021many}, ArxivQA~\cite{li2024multimodal}, Vizcap~\cite{gurari2020captioning}, IconQA~\cite{lu2021iconqa}, CLEVR-Math (CLEVR)~\cite{lindstrom2022clevr}, and Flickr30k~\cite{plummer2015flickr30k}.
CoIN comprises eight sequential tasks: ScienceQA~\cite{lu2022learn}, ImageNet~\cite{deng2009imagenet}, VizWiz~\cite{gurari2018vizwiz}, REC~\cite{kazemzadeh2014referitgame}, TextVQA~\cite{singh2019towards}, GQA~\cite{hudson2019gqa}, VQAv2~\cite{goyal2017making}, and OCR-VQA~\cite{mishra2019ocr}.

\begin{table*}[t]
\centering
\caption{Performance comparison on CoIN. Avg averages each task over the stages
from its introduction onward; Last reports final task scores.
Bold and underlined values denote the best and second-best continual-learning results within each block.}
\label{tab:coin_results}
\small
\setlength{\tabcolsep}{0pt}
\renewcommand{\arraystretch}{1.08}
\setlength{\aboverulesep}{3pt}
\setlength{\belowrulesep}{3pt}
\begin{tabular*}{\textwidth}{@{}
>{\centering\arraybackslash}m{14pt}
>{\raggedright\arraybackslash}m{80pt}
*{9}{>{\centering\arraybackslash}m{\dimexpr(\textwidth-94pt)/9\relax}}@{}}
\toprule
 & \textbf{Method} & \textbf{ScienceQA} & \textbf{ImageNet} & \textbf{VizWiz}
 & \textbf{REC} & \textbf{TextVQA} & \textbf{GQA} & \textbf{VQAv2}
 & \textbf{OCR-VQA} & \textbf{Average}$\uparrow$ \\
\midrule
 & Zero-shot & 69.79 & 9.93 & 45.50 & 58.47 & 57.75 & 60.77 & 66.50 & 64.93 & 54.21 \\
\midrule
\multirow{8}{*}{\rotatebox{90}{\textbf{Avg}}} & FineTune~\cite{hu2022lora} & 64.22 & 40.13 & 43.87 & 38.32 & 55.04 & 55.89 & 60.61 & 64.78 & 52.86 \\
 & LwF~\cite{li2016learning} & 65.20 & 40.63 & 43.22 & 40.05 & 56.23 & 54.67 & 60.64 & 65.12 & 53.22 \\
 & EWC~\cite{kirkpatrick2017overcoming} & 65.11 & 40.89 & 44.09 & 39.67 & 54.92 & 56.03 & 61.12 & 64.55 & 53.30 \\
 & L2P~\cite{wang2022learning} & 70.52 & 26.89 & 45.53 & 45.21 & 56.84 & 59.03 & 63.52 & 64.11 & 53.96 \\
 & O-LoRA~\cite{wang2023orthogonal} & 73.32 & 68.37 & 50.26 & \underline{61.12} & \underline{57.75} & 60.96 & 65.71 & 63.31 & 62.60 \\
 & MoELoRA~\cite{chen2024coin} & 68.38 & 48.50 & 44.22 & 40.23 & 55.62 & 57.04 & 62.14 & \textbf{65.75} & 55.24 \\
 & HiDe-LLaVA~\cite{guo2025hide} & \underline{74.92} & \underline{76.72} & \underline{51.24} & \textbf{61.84} & 57.13 & \textbf{62.83} & \underline{68.15} & 64.76 & \underline{64.70} \\
\cmidrule{2-11}
 & \textbf{CIRA-MoE} & \textbf{82.47} & \textbf{93.19} & \textbf{59.28} & 60.18 & \textbf{63.87} & \underline{61.71} & \textbf{69.79} & \underline{65.51} & \textbf{69.50} \\
\midrule
\multirow{10}{*}{\rotatebox{90}{\textbf{Last}}} & FineTune~\cite{hu2022lora} & 57.43 & 28.90 & 41.88 & 30.05 & 51.39 & 50.76 & 53.28 & 64.78 & 47.31 \\
 & LwF~\cite{li2016learning} & 60.71 & 30.58 & 41.49 & 36.01 & 52.80 & 47.07 & 53.43 & 65.12 & 48.40 \\
 & EWC~\cite{kirkpatrick2017overcoming} & 59.75 & 31.88 & 42.26 & 34.96 & 51.06 & 51.84 & 55.30 & 64.55 & 48.95 \\
 & L2P~\cite{wang2022learning} & 70.21 & 23.31 & 44.21 & 43.76 & 56.25 & 58.46 & 62.32 & 64.11 & 52.83 \\
 & O-LoRA~\cite{wang2023orthogonal} & 72.56 & 62.84 & 48.43 & 58.97 & 57.66 & 59.14 & 63.21 & 63.31 & 60.77 \\
 & MoELoRA~\cite{chen2024coin} & 62.02 & 37.21 & 43.32 & 33.22 & 52.05 & 53.12 & 57.92 & \textbf{65.75} & 50.58 \\
 & HiDe-LLaVA~\cite{guo2025hide} & 73.20 & 69.28 & 50.76 & \underline{59.18} & 56.92 & 61.33 & \textbf{67.12} & 64.76 & 62.82 \\
 & LLaVA-CMoE~\cite{zhao2025llava} & 77.55 & \textbf{94.50} & \underline{55.45} & 23.40 & 58.17 & 48.91 & 56.40 & 59.44 & 59.23 \\
 & SAME~\cite{xie2026same} & \underline{78.35} & 90.21 & 54.13 & \textbf{59.87} & \underline{60.69} & \underline{61.70} & 66.04 & 63.59 & \underline{66.82} \\
\cmidrule{2-11}
 & \textbf{CIRA-MoE} & \textbf{80.02} & \underline{92.36} & \textbf{56.56} & 58.87 & \textbf{61.95} & \textbf{62.05} & \underline{66.90} & \underline{65.51} & \textbf{68.03} \\
\bottomrule
\end{tabular*}
\end{table*}

\paragraph{Compared Methods.}
We compare CIRA-MoE with representative MCIT baselines, including
LoRA fine-tuning (FineTune)~\cite{hu2022lora},
LwF~\cite{li2016learning},
EWC~\cite{kirkpatrick2017overcoming},
L2P~\cite{wang2022learning},
O-LoRA~\cite{wang2023orthogonal},
MoELoRA~\cite{chen2024coin},
CL-MoE~\cite{huai2025cl},
HiDe-LLaVA~\cite{guo2025hide},
SEFE~\cite{chen2025sefe},
LLaVA-CMoE~\cite{zhao2025llava},
and SAME~\cite{xie2026same},
covering knowledge distillation, parameter regularization,
prompt tuning, and routed or expanding LoRA architectures.

\paragraph{Training Details.}
Following prior MCIT settings~\cite{chen2024coin,guo2025hide}, we adopt LLaVA-v1.5-7B~\cite{liu2024improved} with CLIP-L/14-336~\cite{radford2021learning} as the backbone. During training, the pretrained vision encoder and language model are frozen, while current routers, thresholds, and experts are optimized. Each FFN linear projection maintains an independent pool of rank-8 LoRA experts. For each subsequent task, we probe 1\% of its training set and inject up to $K_{\max}=2$ experts per insufficiently covered FFN position. The temperature coefficient $\tau$ and regularization weight $\lambda$ are set to $0.1$ and $5\times10^{-6}$, respectively. Each task is trained for one epoch using a learning rate of $2\times10^{-4}$ and a batch size of 24. All experiments are conducted on four NVIDIA RTX A6000 GPUs.
The first task uses seed experts. Thereafter, probing precedes training,
and historical experts and routers remain frozen. New experts retain
their generation indices for inference. Complete settings and training
and inference algorithms are provided in the supplementary material.

\paragraph{Evaluation Metrics.}
Following prior work~\cite{guo2025hide}, we report \textbf{Avg} and \textbf{Last} across all $N$ tasks. Let $A_{j,i}$ denote the accuracy on task $T_i$ after training on task $T_j$. \textbf{Last} reports the final score $A_{N,i}$, while $\mathrm{Avg}_i=(N-i+1)^{-1}\sum_{j=i}^{N}A_{j,i}$ averages performance on $T_i$ from its introduction onward. The \textbf{Average} column reports the arithmetic mean across all tasks. Ablations and analytical experiments report final task accuracies ($A_{N,i}$) and their mean.

\begin{table}[t]
\centering
\caption{Cumulative ablation of CIRA-MoE on UCIT.}
\label{tab:ablation_ucit}
\small
\setlength{\tabcolsep}{4pt}
\begin{tabular}{@{}l|ccccc@{}}
\toprule
\textbf{Datasets} 
& \textbf{Baseline} 
& \textbf{+EWAR} 
& \textbf{+LWEI} 
& \textbf{+TWRA} 
& \textbf{+ACR} \\
\midrule
ImageNet-R & 51.70 & 55.23 & 57.30 & \textbf{82.17} & \underline{82.10} \\
ArxivQA & 78.17 & 84.40 & 92.50 & \underline{93.33} & \textbf{93.93} \\
Vizcap & 45.10 & 45.60 & 45.95 & \underline{46.35} & \textbf{48.46} \\
IconQA & 68.47 & 69.80 & 69.97 & \textbf{71.27} & \underline{70.10} \\
CLEVR & 48.07 & 58.23 & 56.07 & \underline{61.90} & \textbf{65.73} \\
Flickr30k & 56.96 & \textbf{57.96} & \underline{57.72} & 56.55 & 57.07 \\
\midrule
\textbf{Average}$\uparrow$ 
& 58.08 & 61.87 & 63.25 & \underline{68.60} & \textbf{69.56} \\
\bottomrule
\end{tabular}
\end{table}

\subsection{Benchmark Comparison}
Table~\ref{tab:ucit_results} compares CIRA-MoE with state-of-the-art baselines on UCIT.
CIRA-MoE achieves a mean Avg score of 71.68\% and a mean Last score of 69.56\%, outperforming the runner-up HiDe-LLaVA (68.94\% Avg) by 2.74 percentage points, SEFE (66.54\% Last) by 3.02 percentage points, and CL-MoE (58.49\% Last) by 11.07 percentage points.
Specifically, CIRA-MoE delivers marked gains on reasoning-heavy tasks (IconQA and CLEVR) while maintaining the highest final retention on the earliest tasks (ImageNet-R and ArxivQA), confirming superior adaptation and knowledge preservation.
On the CoIN benchmark (Table~\ref{tab:coin_results}), CIRA-MoE attains 69.50\% in Avg (+4.80 percentage points over HiDe-LLaVA) and 68.03\% in Last, exceeding recent MoE methods SAME and LLaVA-CMoE by 1.21 and 8.80 percentage points, respectively.
Whereas LLaVA-CMoE degrades sharply on intermediate tasks such as REC, CIRA-MoE ranks first or second on seven of the eight tasks in the Last block, demonstrating balanced stability and plasticity across diverse vision--language domains.

\begin{figure}[t]
\centering
\includegraphics[width=\linewidth]{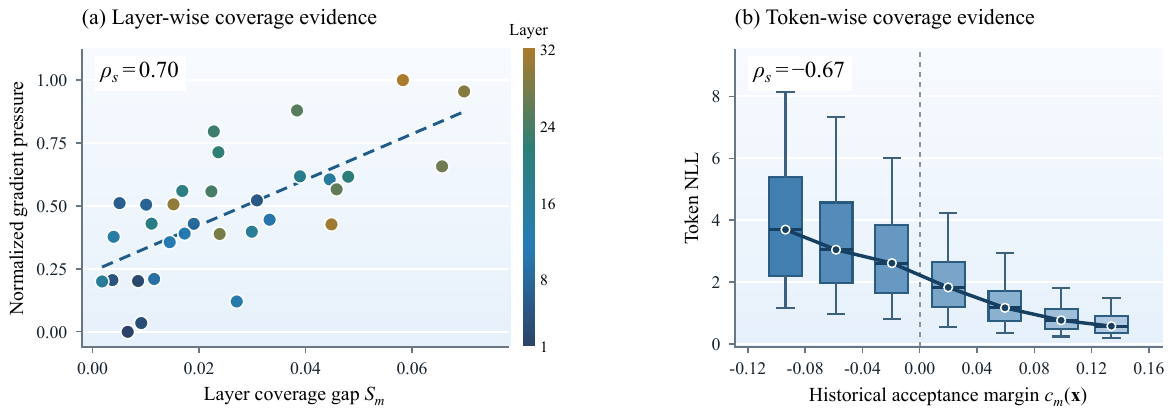}
\caption{Historical coverage diagnostics.
(a) Layer-wise coverage gap $S_m$ versus normalized gradient pressure.
(b) Signed historical acceptance margin $c_m(\mathbf{x})$
versus answer-token NLL.}
\label{fig:coverage_validity}
\end{figure}

\begin{figure*}[t]
    \centering
    \includegraphics[width=1.0\linewidth]{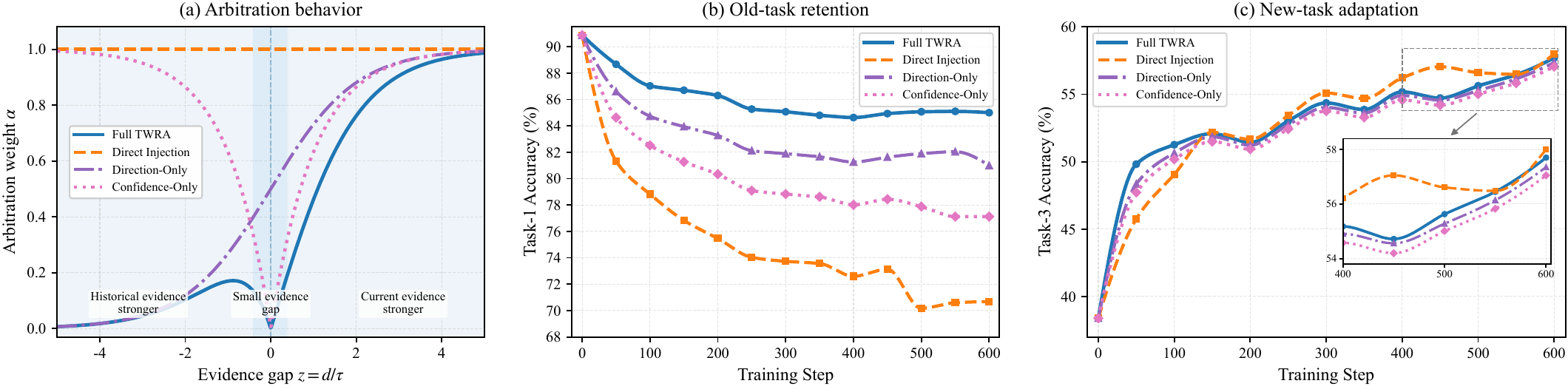}
    \caption{Impact of TWRA.
    Full TWRA substantially improves old-task retention while maintaining new-task performance comparable to Direct Injection. The partial variants verify the complementary effects of the directional and confidence terms.}
    \label{fig:gate_analysis}
\end{figure*}

\subsection{Ablation Study}
Table~\ref{tab:ablation_ucit} reports cumulative ablations on UCIT.
All configurations start with one seed expert per FFN position.
The incremental MoE \textbf{Baseline} uses fixed top-1 routing and uniformly adds two experts per position for each subsequent task.
Introducing \textbf{EWAR} raises average final accuracy from 58.08\% to 61.87\% by enabling flexible multi-expert reuse through threshold-based acceptance.
Adding \textbf{LWEI} further improves accuracy to 63.25\% through selective injection ($\le 2$ experts per position), with a notable improvement on ArxivQA.
Incorporating \textbf{TWRA} produces the largest incremental gain of 5.35 percentage points, reaching 68.60\%, with the strongest improvement on the earliest task, ImageNet-R, indicating better preservation of prior knowledge.
Finally, \textbf{ACR} increases accuracy to 69.56\%, with further gains on reasoning and captioning tasks such as CLEVR and Vizcap, supporting the benefits of suppressing redundant activation and encouraging expert specialization.

\begin{figure}
    \centering
    \includegraphics[width=\linewidth]{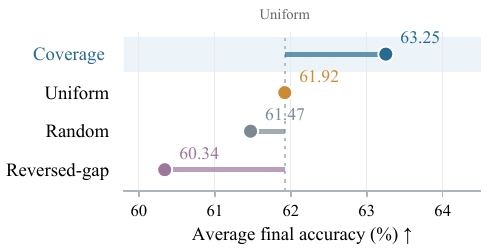}
    \caption{Comparison of layer-wise expert allocation strategies on UCIT.
    EWAR is fixed, TWRA and ACR are disabled, and added parameter budgets are matched at each stage.
    The dashed line marks the Uniform baseline.}
    \label{fig:lwei_allocation}
\end{figure}

\subsection{Further Analysis}
\paragraph{Validity of Coverage Evidence.}
We first examine whether our historical acceptance evidence faithfully captures the ground-truth adaptation demands observed in Fig.~\ref{fig:motivation} at both the layer and token levels (Fig.~\ref{fig:coverage_validity}).
As shown in Fig.~\ref{fig:coverage_validity}(a), the layer-wise coverage gap $S_m$ exhibits a strong positive correlation with normalized gradient pressure (Spearman correlation $\rho_s=0.70$).
This indicates that layers with weaker historical coverage experience higher loss sensitivity to parameter updates on the incoming task, justifying LWEI's prioritization of high-$S_m$ layers for capacity expansion.
At the token level, Fig.~\ref{fig:coverage_validity}(b) shows that the signed historical acceptance margin $c_m(\mathbf{x})$ correlates negatively with target answer-token NLL ($\rho_s=-0.67$).
Tokens accepted with positive margins ($c_m(\mathbf{x})>0$) consistently incur low NLL and are already well handled by frozen historical experts, whereas tokens falling near or below the rejection boundary ($c_m(\mathbf{x})\le 0$) suffer substantially higher prediction difficulty---supporting TWRA's design of attenuating new-expert residuals on well-covered tokens while intervening where historical coverage fails.
Together, these results confirm that forward acceptance margins provide a reliable, unified proxy for guiding both layer-wise expert injection and token-wise residual arbitration.

\begin{table}[t]
\centering
\caption{Gate controls on UCIT without LWEI or ACR.
Average final accuracy (\%) is reported.}
\label{tab:token_weight_alignment}
\normalsize
\renewcommand{\arraystretch}{1.10}
\setlength{\aboverulesep}{2pt}
\setlength{\belowrulesep}{2pt}
\begin{tabular*}{0.76\columnwidth}{
@{\hspace{4pt}\extracolsep{\fill}}lc@{\hspace{4pt}}}
\toprule
\textbf{Gate} & \textbf{Average}$\uparrow$ \\
\midrule
Direct Injection & 61.87 \\
Shuffled Gate & 60.24 \\
Mean Gate & 64.42 \\
\textbf{TWRA} & \textbf{67.18} \\
\bottomrule
\end{tabular*}
\end{table}

\paragraph{Impact of Layer-Wise Expert Injection.}
We compare layer-wise allocation strategies under an identical stage-wise parameter budget, keeping EWAR fixed while disabling TWRA and ACR.
Specifically, \textit{Uniform} distributes new experts evenly across FFN positions, \textit{Random} samples positions at random, and \textit{Reversed-gap} prioritizes positions with smaller coverage gaps $S_m$.
As shown in Fig.~\ref{fig:lwei_allocation}, coverage-guided LWEI (\textit{Coverage}) achieves 63.25\%, outperforming \textit{Uniform}, \textit{Random}, and \textit{Reversed-gap} by 1.33, 1.78, and 2.91 percentage points, respectively.
Notably, inverting the allocation priority (\textit{Reversed-gap}) drops accuracy even below \textit{Uniform} and \textit{Random}, confirming that starving high-$S_m$ layers while over-parameterizing well-covered layers severely hampers continual adaptation.


\paragraph{Impact of Token-Wise Residual Arbitration.}
Fig.~\ref{fig:gate_analysis}(a) illustrates the complementary
roles of direction and confidence: direction alone retains
substantial intervention near ties, whereas confidence alone
increases intervention even under historical dominance.
As shown in Fig.~\ref{fig:gate_analysis}(b)--(c), TWRA
improves Task~1 accuracy during Task~3 training from
$70.67\%$ with \textit{Direct Injection} to $85.23\%$,
while maintaining comparable Task~3 accuracy
($57.87\%$ vs.\ $57.70\%$).
\textit{Direction-Only} and \textit{Confidence-Only}
also improve retention, achieving $80.99\%$ and $77.12\%$
on Task~1, respectively, but neither improves Task~3 adaptation.
These results support the complementary roles of direction
and confidence in reducing historical interference while
preserving current-task learning.

To isolate the effect of token-specific weighting, we disable
LWEI and ACR and compare TWRA with \textit{Mean Gate} and
\textit{Shuffled Gate}.
Within each routed position and task group, these variants
respectively broadcast the batch-mean TWRA weight or permute
the weights among non-padding tokens with nonempty accepted
expert sets.
As shown in Table~\ref{tab:token_weight_alignment}, TWRA
achieves $67.18\%$, exceeding \textit{Direct Injection},
\textit{Mean Gate}, and \textit{Shuffled Gate} by $5.31$,
$2.76$, and $6.94$ percentage points, respectively.
Mean weighting improves over direct injection, suggesting
that uniform attenuation is beneficial but insufficient.
Shuffling performs worse despite preserving the weight
distribution, highlighting the importance of matching
residual strength to each token's evidence.

\paragraph{Analysis of Different Task Orders.}
Table~\ref{tab:task_order_robustness} evaluates CIRA-MoE across three task permutations on UCIT.
The average final accuracy remains highly stable between $69.37\%$ and $69.97\%$ (a spread of only $0.60$ percentage points), demonstrating strong robustness to task ordering.

\begin{table}[t]
\centering
\caption{Task-order robustness on UCIT. Final accuracy (\%) is reported in training order. R: ImageNet-R; A: ArxivQA; V: Vizcap; I: IconQA; C: CLEVR; F: Flickr30k.}
\label{tab:task_order_robustness}
\footnotesize
\setlength{\tabcolsep}{2pt}
\renewcommand{\arraystretch}{1.06}
\setlength{\aboverulesep}{2pt}
\setlength{\belowrulesep}{2pt}
\begin{tabular*}{\columnwidth}{@{\extracolsep{\fill}}l*{7}{c}@{}}
\toprule
\textit{Order 1} & \textbf{R} & \textbf{A} & \textbf{V} & \textbf{I} & \textbf{C} & \textbf{F} & \textbf{Average}$\uparrow$ \\
\cmidrule{2-8}
\texttt{RAVICF} & 82.10 & 93.93 & 48.46 & 70.10 & 65.73 & 57.07 & \textbf{69.56} \\
\midrule
\textit{Order 2} & \textbf{A} & \textbf{I} & \textbf{R} & \textbf{F} & \textbf{C} & \textbf{V} & \textbf{Average}$\uparrow$ \\
\cmidrule{2-8}
\texttt{AIRFCV} & 86.50 & 66.37 & 82.90 & 52.82 & 69.93 & 57.69 & \textbf{69.37} \\
\midrule
\textit{Order 3} & \textbf{I} & \textbf{F} & \textbf{R} & \textbf{C} & \textbf{A} & \textbf{V} & \textbf{Average}$\uparrow$ \\
\cmidrule{2-8}
\texttt{IFRCAV} & 63.40 & 54.90 & 83.74 & 68.73 & 92.72 & 56.33 & \textbf{69.97} \\
\bottomrule
\end{tabular*}
\end{table}

\begin{figure}
    \centering
    \includegraphics[width=\linewidth]{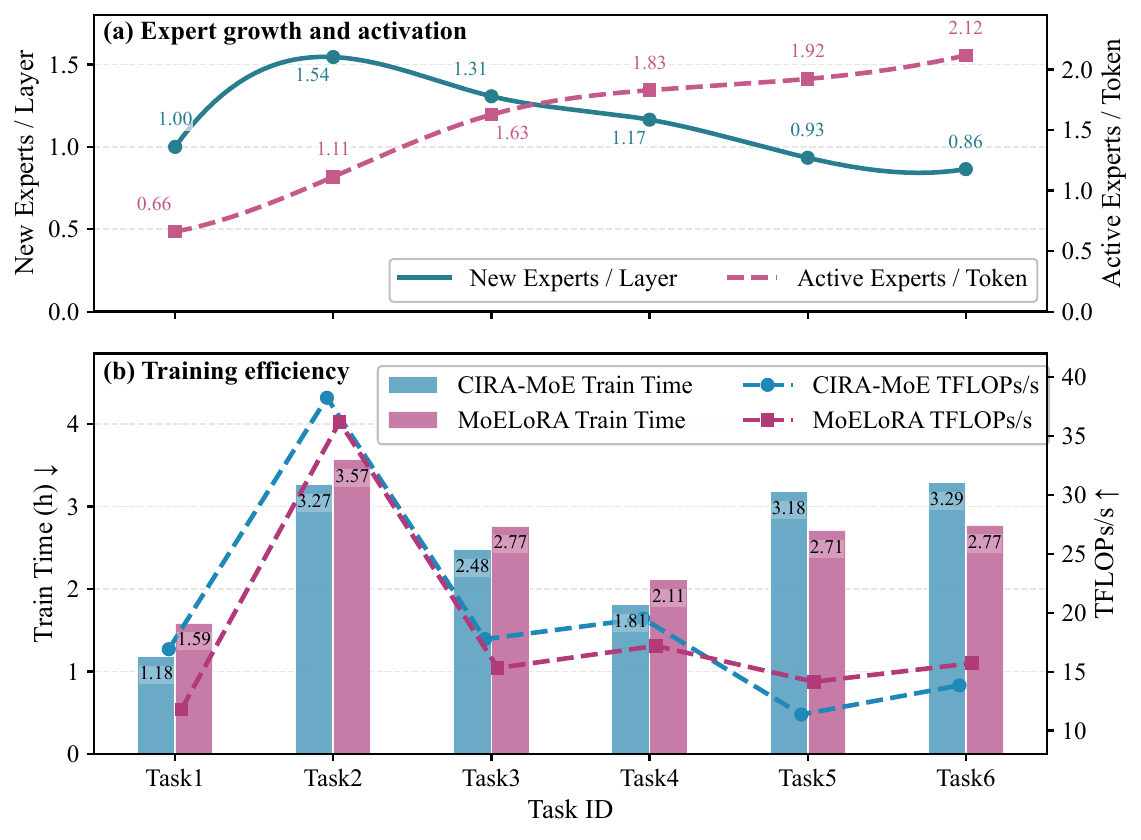}
    \caption{Expert growth and efficiency of CIRA-MoE.}
    \label{fig:model_growth_efficiency}
\end{figure}

\paragraph{Expert Growth and Efficiency.}
Fig.~\ref{fig:model_growth_efficiency}(a) averages expert additions
across routed FFN positions and activations across positions and tokens.
New experts per position decrease by $44.2\%$ from $1.54$ (Task~2)
to $0.86$ (Task~6), while active experts increase from $0.66$ to $2.12$.
Thus, later stages add fewer experts while activating more of the
accumulated pool. Fig.~\ref{fig:model_growth_efficiency}(b) reports
$2.54$~h of training per task, compared with $2.59$~h for MoELoRA,
and achieved computational throughput of $19.61$ versus
$18.40$~TFLOPs/s.

\begin{table}[t]
\centering
\caption{Hyperparameter sensitivity of CIRA-MoE on UCIT.
Bold values denote the default settings and corresponding results.}
\label{tab:param_sensitivity}
\small
\setlength{\tabcolsep}{2pt}
\begin{tabular*}{\columnwidth}{@{\extracolsep{\fill}}ccc@{}}
\toprule
\textbf{Parameter} & \textbf{Settings} & \textbf{Average}$\uparrow$ \\
\midrule
$\tau$
& $0.05,\,\mathbf{0.10},\,0.50$
& $68.31,\,\mathbf{69.56},\,68.92$ \\
$\lambda$
& $5{\times}10^{-4},\,\mathbf{5{\times}10^{-6}},\,1{\times}10^{-6}$
& $68.45,\,\mathbf{69.56},\,68.14$ \\
$\rho$
& $0.1\%,\,\mathbf{1\%},\,5\%$
& $68.52,\,\mathbf{69.56},\,69.51$ \\
\bottomrule
\end{tabular*}
\end{table}

\paragraph{Hyperparameter Sensitivity.}
Table~\ref{tab:param_sensitivity} evaluates sensitivity to the arbitration temperature $\tau$, regularization weight $\lambda$, and probing ratio $\rho$.
The intermediate $\tau$ and $\lambda$ settings outperform both
tested extremes. Increasing $\rho$ from $0.1\%$ to $1\%$ raises
performance by $1.04$ percentage points. A $5\%$ probe yields
$69.51\%$, comparable to $69.56\%$ at $1\%$, supporting the
smaller probe for expert allocation.

\section{Conclusion}
We identify layer- and token-wise heterogeneity in the adaptation demands of MoE-based MCIT.
Motivated by these findings, we propose CIRA-MoE, which leverages expert acceptance margins as a unified signal to estimate historical coverage, coordinating Layer-Wise Expert Injection (LWEI), Token-Wise Residual Arbitration (TWRA), and Arbitration-Consistent Regularization (ACR).
CIRA-MoE selectively allocates new capacity to poorly covered layers and regulates current-expert residual contributions according to token-level adaptation needs.
Experiments on UCIT and CoIN demonstrate that CIRA-MoE outperforms state-of-the-art methods in MCIT.


\clearpage
{
    \small
    \bibliographystyle{ieeenat_fullname}
    \bibliography{main}
}


\end{document}